%% file: main.tex
\documentclass{article}

\PassOptionsToPackage{numbers}{natbib}

\usepackage[final]{neurips}
\usepackage{amssymb}

\usepackage{subfig}
\usepackage{graphicx}
\usepackage{tabularx}
\usepackage{soul}
\usepackage{multirow}
\usepackage{hhline}

\usepackage{epstopdf}
\usepackage[utf8]{inputenc}

\usepackage{hyperref}
\usepackage{xstring}
\usepackage{amsmath,amssymb,amsfonts}
\usepackage{algorithmic}
\usepackage{graphicx}
\usepackage{xcolor}
\usepackage{verbatim}
\usepackage{authblk}

\usepackage{amsmath}
\usepackage{langsci-gb4e}

\title{Spoken Dialogue System for Medical Prescription Acquisition on Smartphone: Development, Corpus and Evaluation}

\begin{document}
\author[1,2]{Ali Can Kocabiyikoglu}
\author[1]{François Portet}
\author[2]{Jean-Marc Babouchkine}
\author[3]{Prudence Gibert}
\author[1]{Hervé Blanchon}
\author[4]{Gaëtan Gavazzi}

\affil[1]{Univ. Grenoble Alpes, CNRS, Grenoble INP, LIG, 38000 Grenoble, France}
\affil[2]{Calystene SA,16 Rue Irene Joliot Curie, 38320 Eybens, France}
\affil[3]{CHU Grenoble Alpes, Avenue Maquis-du-Gresivaudan, 38700 La Tronche, France}
\affil[4]{Clinique de medecine geriatrique, CHU Grenoble Alpes, Equipe Grepi, EA 7408, CS 10217, 38700 La Tronche, France}

\maketitle
\begin{abstract}
Hospital information systems (HIS) have become an essential part of healthcare institutions and now incorporate prescribing support software. Prescription support software allows for structured information capture, which improves the safety, appropriateness and efficiency of prescriptions and reduces the number of adverse drug events (ADEs). However, such a system increases the amount of time physicians spend at a computer entering information instead of providing medical care. In addition, any new visiting clinician must learn to manage complex interfaces since each HIS has its own interfaces. In this paper, we present a natural language interface for e-prescribing software in the form of a spoken dialogue system accessible on a smartphone. This system allows prescribers to record their prescriptions verbally, a form of interaction closer to their usual practice. The system extracts the formal representation of the prescription ready to be checked by the prescribing software and uses the dialogue to request mandatory information, correct errors or warn of particular situations. Since, to the best of our knowledge, there is no existing voice-based prescription dialogue system, we present the system developed in a low-resource environment, focusing on dialogue modeling, semantic extraction and data augmentation. The system was evaluated in the wild with 55 participants. This evaluation showed that our system has an average prescription time of 66.15 seconds for physicians and 35.64 seconds for other experts, and a task success rate of 76\% for physicians and 72\% for other experts. All evaluation data were recorded and annotated to form PxCorpus, the first spoken drug prescription corpus that has been made fully available to the community (\url{https://doi.org/10.5281/zenodo.6524162}). 
\end{abstract}

\section{Introduction}\label{sec:intro}


Hospital Information Systems (HIS) have become an essential part of health institutions aiming to provide better healthcare services by improving the organization as well as the traceability and the quality of care~\citep{lau2010review}. HIS makes it possible to have a centralized database and software platform to deal with the discharge of patients, the given treatments, drugs, pharmacy details and so on. One of the tasks HIS is particularly useful for is the support for electronic prescriptions provided by e-Prescribing software. Using this kind of software, physicians are assisted during the computerized entry of prescriptions. Today, there is a consensus on its effectiveness in preventing adverse drug events (ADEs)\citep{mille2005evaluation}.

ADEs often result of medication errors that are a major concern in health care. Indeed, in the USA, medication errors are experienced by 1.5 million patients per year\citep{aspden2006committee}. In France, a similar study shows that serious adverse events are estimated to be around 350 000 to 460 000 cases a year\citep{haury2005evenements}. Studies of past ADEs have shown that most of them are avoidable\citep{barker2002medication,kaushal2001medication} especially using HIS and information technology\citep{agrawal2009medication}. Using an integrated software for e-prescriptions allows structured information entry enabling better security, adequacy, and efficiency of prescriptions. However, HIS and e-Prescription software are fully efficient when all the medical information is entered digitally into the system. As a consequence, this has increased the amount of time nurses and physicians spend in front of a computer entering information instead of providing medical care \cite{devine2010electronic,10.1136/amiajnl-2012-001414}.

\begin{figure}[bht]
    \centering
    \begin{tabular}{c|c}
         \includegraphics[width=0.55\textwidth]{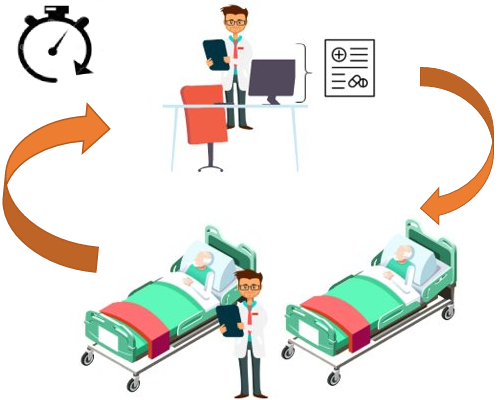}&\includegraphics[width=0.35\textwidth]{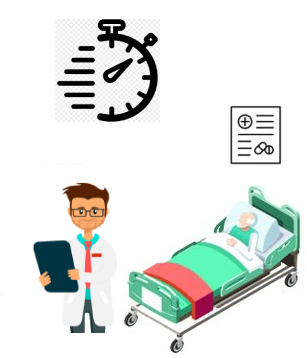}  \\
    \end{tabular}
    \caption{Prescription process in a hospital facility. Left: usual practice where the e-prescription keyboard interface is far from the bedside; Right: our system where e-prescription is available at the point of care through natural language.}
    \label{fig:medecinDeplace}
\end{figure}

In this paper, we present a natural language interface to the e-Prescription software in the form of a spoken dialogue system.
This interface would enable practitioners to record their prescriptions orally or through free text, a form of interaction closer to their usual practice. It would also enable practitioners to use natural language interaction so that they do not have to learn new complex software interfaces depending on different HIS. We designed the application so that clinicians could use their own smartphones. In this way, they can get quickly familiar with the application while enabling faster identification than standard login. Mobile interfaces would also enable clinicians to prescribe at the point of care which would save time and would enable better mobility\cite{altieri}. Indeed, in a healthcare facility, physicians move around the hospital, visiting patients' rooms and performing diagnosis, follow-ups and re-evaluations. Figure~\ref{fig:medecinDeplace} (left) describe this process. After visiting patients, physicians enter the drug prescriptions by using e-Prescription software, often only accessible from a desktop computer. However, at the point of care prescribers do not always have the time or the possibility to access the fixed stations to enter their prescriptions. It is therefore very frequent that the data entry takes place on paper, written by hand or even orally, which can be a source of avoidable ADEs and which make the traceability of care more difficult. With a spoken dialogue system on a mobile phone (Fig.~\ref{fig:medecinDeplace} right), the prescription can be performed at the point of care without complex interfaces. 

In addition, a dialogue-based prescribing system could inform the clinician for drugs that are not available in the pharmacy and inform the practitioner of potential adverse drug reactions via dialogue. It could also provide personalized patient-related information (e.g. about allergies) so that the practitioner can adapt the prescription in real time.

\textbf{Paper Contributions.}  In this paper we present the full approach that enabled the creation of a medical drug prescription dialogue system. In particular, we describe:
\begin{itemize}
\item the methodology to build a complete spoken drug prescription dialogue system with low resource;
\item the data collection and augmentation methods to train state-of-the-art dialogue modules (NLU, drug disambiguation, dialogue policy) in a low resource setting;
\item a detailed evaluation of the system with 55 participants (physicians, drug experts and naive speakers) in wild conditions;
\item PxCorpus a drug prescription dataset comprising around 4h of speech recordings acquired during the evaluation and humanly annotated which have been made freely available with a Creative common license. This corpus is intended to support design and evaluation of prescription dialogue systems.
\end{itemize}
This paper builds on some previous work \citep{kocabiyikoglu2019towards,kocabiyikoglu2022}. In \citep{kocabiyikoglu2019towards}, some preliminary natural language understanding (NLU) methods were presented. In this article, the NLU models were Transformer pre-trained models which largely outperform the previous NLU models in \citep{kocabiyikoglu2019towards}. In \citep{kocabiyikoglu2022},  the acquisition of a spoken drug prescription corpus for spoken language understanding (PxSLU) was described. In this paper, we extend this work by presenting the full dialogue system, the data augmentation strategy and more details about the results of the evaluation. 

\textbf{Plan.} Section \ref{sec:sota} briefly introduces related work on dialogue systems used in healthcare and present related shared tasks and challenges as well as drug prescription datasets. Section \ref{sec:method} describes the method that we used for creating a mobile goal-oriented dialogue system for prescribing medicine. This is followed by the section \ref{sec:dialogue} where we detail the modules of the dialogue system. Section \ref{sec:results} presents the medical prescription data that we acquired through dialogue and reports our findings on natural language understanding (NLU) results and characteristics and metrics of the collected dialogues. Finally, section \ref{sec:conclusion} concludes this paper with a brief discussion and gives directions for future research.

\section{Related work}\label{sec:sota}

Systematic reviews on health dialogue systems show that the use of dialogue systems in healthcare has a long history and represents a rapidly growing field with a large number of active researchers and increasing number of systems built and deployed in the field\citep{bickmore2006health, kearns2019systematic}. 
Health dialogue systems focus mostly on the preventive healthcare of patients, especially in the context of mental health~\citep{fitzpatrick2017delivering,miner2016smartphone, hudlicka2013virtual} but other application areas exist such as preventive diagnostics~\citep{beveridge2006automatic, philip2014could}, ambulatory monitoring~\citep{giorgino2005automated}, health data collection~\citep{black2005appraisal}, etc. Most of these dialogue systems focus on  dialogue with patients, however there are systems that are designed for medical experts. For example, \cite{llanos2015description, campillos2020designing} presents a dialogue system that simulates a patient during a medical consultation for educational purposes. 

Despite the high interest in medical entity extraction task from health records, research on extraction of medical drug prescriptions are not numerous. For example, \cite{tao2018fable} propose a medical prescription extraction system based on a semi-supervised method from I2B2-2009 medication extraction dataset\cite{uzuner2010community, uzuner2010extracting}. However, information extraction tasks that are used in i2b2 challenges focus on information extraction from long textual documents such as discharge summaries or medical records while our focus is on short spoken utterances. 

\subsection{Dialogue based prescription systems}

Processing of short utterances for spoken medical prescriptions can be found in a few papers or existing systems. The first one is  FreePharma{\footnotesize\textcopyright}~ cited in a chapter in 2006~\cite{dossantos2006} which is described as a system capable of medical prescriptions from speech using PDAs. However, no technical details are provided and the reference points to a dead link.  Another related work is the European Mobi-Dev project\cite{altieri} which aimed to provide the next generation of mobile devices for clinicians at the point of care. However, we have not found any scientific publications related to this project. One of the more recent works published on this task was \cite{ikhu2010voice} proposed in 2010. The authors present a dialogue system for spoken medical drug prescriptions in order to avoid errors and potentially detect ADEs and drug interactions based on VoiceXML\cite{lucas2000voicexml} uttered in a specific order through a phone call. However, the authors do not present any evaluation of the system or any concrete development.

More recently, some works related to spoken prescriptions have been proposed~\citep{mahatpure2019electronic, sanjeev2021advanced, babu2021voice}. These work remain very focused on the software implementation of a voice dictation of medical prescriptions and their transcriptions for their transmissions in order to avoid misunderstandings related to the reading of prescriptions. Their evaluation on an unspecified population shows that voice dictation of a prescription would be faster than typing on a smartphone keyboard. No semantic or error analysis is mentioned. We can also mention a recent dialogue system (\textit{DocPal}) allowing accessing and modifying medical reports (EHR) orally~\citep{bhatt2021docpal}. The satisfaction questionnaire on this first prototype shows that participants found this voice assistant easy to use, fast and accurate.

This brief overview shows there is an interest in the spoken medical prescription domain. However the lack of scientific publications and patents indicate that these projects have not been pursued.

\subsection{Drug prescription datasets for NLP}

While the area of spoken drug prescription processing has only been sparsely addressed, textual drug prescriptions have received more attention. Indeed, drug prescriptions are found in the two main types of public datasets in the biomedical NLP domain. The first type is the big institutional data warehouses and the second type is the datasets that are collectively built during biomedical challenges and academic datasets for a specific task. 

Most of the biomedical NLP research uses big institutional warehouse datasets such as MIMIC-III~\cite{johnson2016mimic} or AP-HP Health Data Warehouse\footnote{\url{https://eds.aphp.fr/}}. These datasets are distributed as a database with deidentified information and are used for numerous tasks. 

\begin{table}[thb]
    \centering
    {\footnotesize
    \input{./challenges.tex}
    }
    \caption{Shared tasks and challenges for medication and ADE extraction}
    \label{tab:challenges}
\end{table}

However, these datasets are not annotated for specific tasks. That is why there has been a considerable effort in creating challenges for specific tasks. Such challenges involve and stimulate data collection, annotation, evaluation and tools that are open to the scientific community. Table~\ref{tab:challenges} summarizes some challenges that are relevant to information extraction for drugs and ADEs. The most commonly used challenge datasets are I2B2 (Informatics for Integrating Biology and the Bedside), N2C2 (National NLP Clinical Challenges), and SemEval. These datasets are made of texts that are more substantial than simple prescriptions, such as discharge summaries or Electronic Health Records (EHRs).

Regarding drug prescriptions only, some datasets include drug prescriptions written in natural language. For example, the I2B2 challenge dataset contains medical prescriptions mostly in narrative form inside EHRs such as in I2B2 medication extraction challenge\cite{uzuner2010extracting} and Medication and Adverse Drug Events Challenge 2019~\cite{jagannatha2019overview}.  Another source of prescriptions could be the MedDialog dataset \cite{chen2020meddiag}. It is composed of 0.26 million medical consultations in English and 1.1 million in Chinese scraped from online platforms. Each dialogue contains a description of patient’s medical condition, a conversation between a patient and a physician and optionally diagnosis and treatment suggestions. However, the prescription part is not annotated and the dataset is only textual.

Finally, even though related datasets exist in other languages such as QUAERO corpus\cite{neveol14quaero} for Named Entity Recognition (NER) in French, there is a clear lack of drug prescription corpora in language other than English. 
It exists one corpora composed of textual medical prescription in French from \cite{grouin2011acces}, however, the authors did not give us access to the corpus since it is composed of proprietary data.

To summarize this short overview of the state-of-the-art, there is, to the best of our knowledge, no existing working systems to acquire spoken drug prescriptions on mobile phone connected to a HIS. Furthermore, there is no available dataset of spoken drug prescriptions expressed in a natural way in any language.
This paper addresses these two shortcomings by describing the method to build a complete spoken drug prescription system from scratch, its implementation and the results of its evaluation with naive participants and physicians. Furthermore, we provide the community with the first corpus of spoken drug prescriptions in French fully annotated with NLU and dialogue labels to encourage reproducibility \cite{kocabiyikoglu2022}. 




\section{Method}\label{sec:method}



We approach the spoken medical prescription understanding problem as a dialogue task in which the utterance initiated by the user must be understood, disambiguated and complemented trough goal oriented dialogue. Since many pieces of information may not be present in oral prescriptions or wrongly recognized, a dialogue system makes it possible to recover errors and to seek missing information. This allows as well to validate information and make sure that the prescription is correct because of the multiple steps of implicit and explicit confirmation. Figure~\ref{fig:dialogueSteps} gives an example dialogue with all the different steps of the system.

\begin{figure}[!htb]
\begin{exe}
    \ex \textbf{Dialogue initiated:} \\
    \emph{Prescriber}: Ofloxacine 200 mg 2 injections per day\label{ex:1}
    \ex \textbf{Semantic Information Extraction:\\}
    \glll Ofloxacine 200 mg 2 injections per day\\
    INN d-dos-val d-dos-up dos-val dos-uf O O\\
    \vphantom{} \vphantom{} \vphantom{} \vphantom{} \vphantom{}\\
    \vspace{-0.5cm} 
    \label{ex:2}
    \ex \textbf{Disambiguation and Information Filling:\\} (OFLOXACINE 200 mg/40 ml, solution for intravenous infusion, intravenous route)
    \label{ex:3}
    \ex \textbf{Requesting precision from the prescriber:\\} \emph{System}: Can you please specify a duration for this prescription?\\
    \emph{Prescriber}: For 7 days
    \label{ex:4}
    \ex \textbf{Proposition of a structured prescription:\\} Ofloxacine 200 mg/40 ml, injectable solution, route of administration is intravenous, 2 injections per day everyday for 1 week. Do you confirm this prescription?
    \label{ex:5}
    \ex \textbf{Checking for drug interactions and patient history:\\}
    \emph{System}: Contraindication detected, the patient has the following pre-existing medical conditions...\\
    \emph{Prescriber}: Cancel
    \label{ex:6}
\end{exe}
    \caption{Example dialogue}
    \label{fig:dialogueSteps}
\end{figure}

In step \textbf{(1)}, the dialogue is initiated by the prescriber with a spoken utterance. The spoken utterance is first transcribed into text by an ASR module. Step \textbf{(2)}, NLU module extracts the intent of the user and the semantic labels related to the prescription information using a \emph{slot-filling}. For instance, Ofloxacine is recognized as an International Non-proprietary Name (INN) as well as various information about the quantity to inject. 
At step \textbf{(3)}, the name of the drug is retrieved from an official drug database. Since several forms of administration and dosage exist, we use all contextual information to restrict the number of potential candidates. Here, despite the route of administration is not explicitly stated, the information about the dose and the injection can be used to retrieve only one candidate. At this step, if several candidates are matched, a list of drugs is proposed to the prescriber who must choose one or restart the prescription. 
Step \textbf{(4)} aims at requesting further information from the prescriber to comply with the latest e-prescribing regulations. During this step, the prescriber can also add non-mandatory information (complex constraints).  Once the entire requested information is provided, the prescription is summarized to the prescriber to be confirmed explicitly as shown in \textbf{(5)}. Finally, the validation process of the e-prescribing software could warn the practitioner for contraindications and patient background as in step \textbf{(6)}.

To design the system, we considered a goal-oriented dialogue (i.e. complete a drug prescription) using a slot-filling approach (the mandatory and optional elements of a prescription). This system must be connected to an HIS in order to gather information about the patient (e.g. contraindication) and the hospital and to enter the prescription of the patient. Such system should rely on medical knowledge bases and several NLP modules (e.g. ASR, NLU, dialogue). To achieve this, several challenges needed to be addressed.
\begin{itemize}
    \item Although, drug prescription is a concept present in any information systems, there is no freely available taxonomy fit for slot-filling that can be directly used to design the semantic of the dialogue system
    \item Since spoken prescriptions in practice exist only through dictation, there is no actual example of spoken drug prescription dialogue to base our design on. 
    \item There is no dataset of drug prescriptions in French and to the best of our knowledge no spoken drug prescription dataset available in any language. Our system thus had to be designed in a low resource setting.
    \item This work was performed during the Covid-19 pandemic making it hard to access physicians. 
\end{itemize}

\begin{figure}[htbp]
    \centering
    \includegraphics[width=0.45\textwidth]{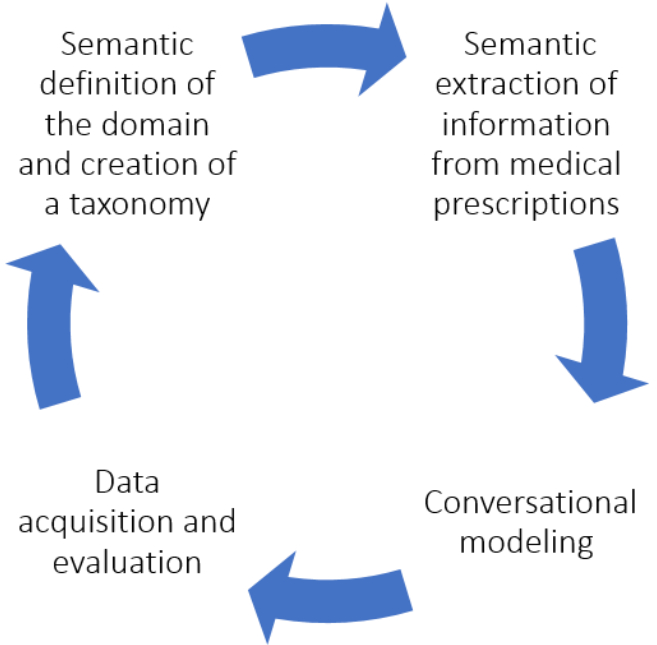}
    \caption{Steps of the iterative method}
    \label{fig:iterativeApproach}
\end{figure}

To address all these challenges, we used an iterative approach. Figure~\ref{fig:iterativeApproach} shows the steps that we have followed. Using, the information collected by an existing e-prescription system, we first defined the semantic space of the information to be collected. Using, this space, we collected an initial set of written prescriptions that we annotated using this semantic space. By using this initial set of data, we crossed these exemplars with the mandatory and optional slots to design credible scenarios. This data and scenarios were used to design several versions of the system, which were analyzed and corrected iterative, to generate new data, improve scenarios and so on. To validate each step we relied on two main experts (co-authors of this paper) who were respectively a pharmacist and a physician expert in e-prescription. The different versions of the system were also tested with experts in prescription who were not part of the project.     

After two rounds of the steps shown in \ref{fig:iterativeApproach}, we have integrated the system with a professional e-Prescription software and tested it with physicians, biologists, pharmacists and biomedical engineers to validate our approach.  The following subsections details how we designed the semantics of the system and how we dealt with the low resource by data augmentation. While the remaining of the paper details the dialogue design and the evaluation.

\subsection{Semantics of Medical Prescriptions}\label{subsec:semantics}

A goal-oriented modular dialogue system depends on an explicit definition of a domain, semantic slots and normalised values. This is performed using a NLU method using 
the \emph{slot-filling} approach. Slot-filling consists in extracting the overall \emph{intent} of an utterance and identifying the most important elements called \emph{slots}. The intent reflects the intention of the speaker (such as greeting, requesting, etc.) while the slots can be defined as the entities and relations in the utterance which are relevant for the given task~\cite{tur:2011}.

In the biomedical NLP domain, there are lots of work around information representation using knowledge bases or ontologies related to the semantic representation of drugs and chemicals. For example LinKBase\textsuperscript{\textregistered}~\cite{van2006linkbase} is an ontology which was specifically designed for NLU applications and includes a lexicon of 1.5 million terms for the biomedical domain. Specifically for medical prescriptions, we can cite \emph{PDRO} a Canadian bilingual (english-french) ontology~\cite{Ethier2016ImprovingTS} which aims to model entities related to a prescription to avoid ambiguities (repetition, interruption, etc.). In the USA, there is also the drug nomenclature developed by the National Library of Medicine (RxNorm)~\cite{liu2005rxnorm} which is widely used by the community. Moreover, in the Web Semantics domain, there are multiple ontologies developed for representing medical concepts related to prescriptions \cite{grando2012ontological,khalili2013semantic}.

The aforementioned ontologies have proven to be useful but not sufficient to represent all of the semantic information in the medical prescriptions domain. This is because prescriptions do not only include administrative information but also remarks and indications for the patient. Especially, when the prescription is in a spoken form, prescriptions can include information in free-form such as ``take on empty stomach'' or ``make sure to drink a full glass of water after\dots''. Since our goal was to provide a natural language interface to an e-prescription software, we developed a taxonomy inspired from the drug ontologies cited above and e-Prescription regulations in France~\cite{LAP2016}. These regulations are given by the French National Authority for Health (HAS) in detailed reports and regularly reviewed. The certified e-prescription software we used was based on these regulations. Regarding semantic slots, some of them were inspired directly from these regulations such as minimum gaps between two administrations or maximum dose given in a time-frame when a drug administration is conditional (e.g. take 1 tablets, in case of a severe headache, take 2 maximum per 24h). Even though the prescription examples that we have studied showed very few examples written in this way, we included these possibilities in our semantic definition and reviewed it with medical experts at the beginning of the iterative method described Figure~\ref{fig:iterativeApproach}.

\begin{figure}[htbp]
\centering
    \includegraphics[angle =90,width=0.85\textwidth]{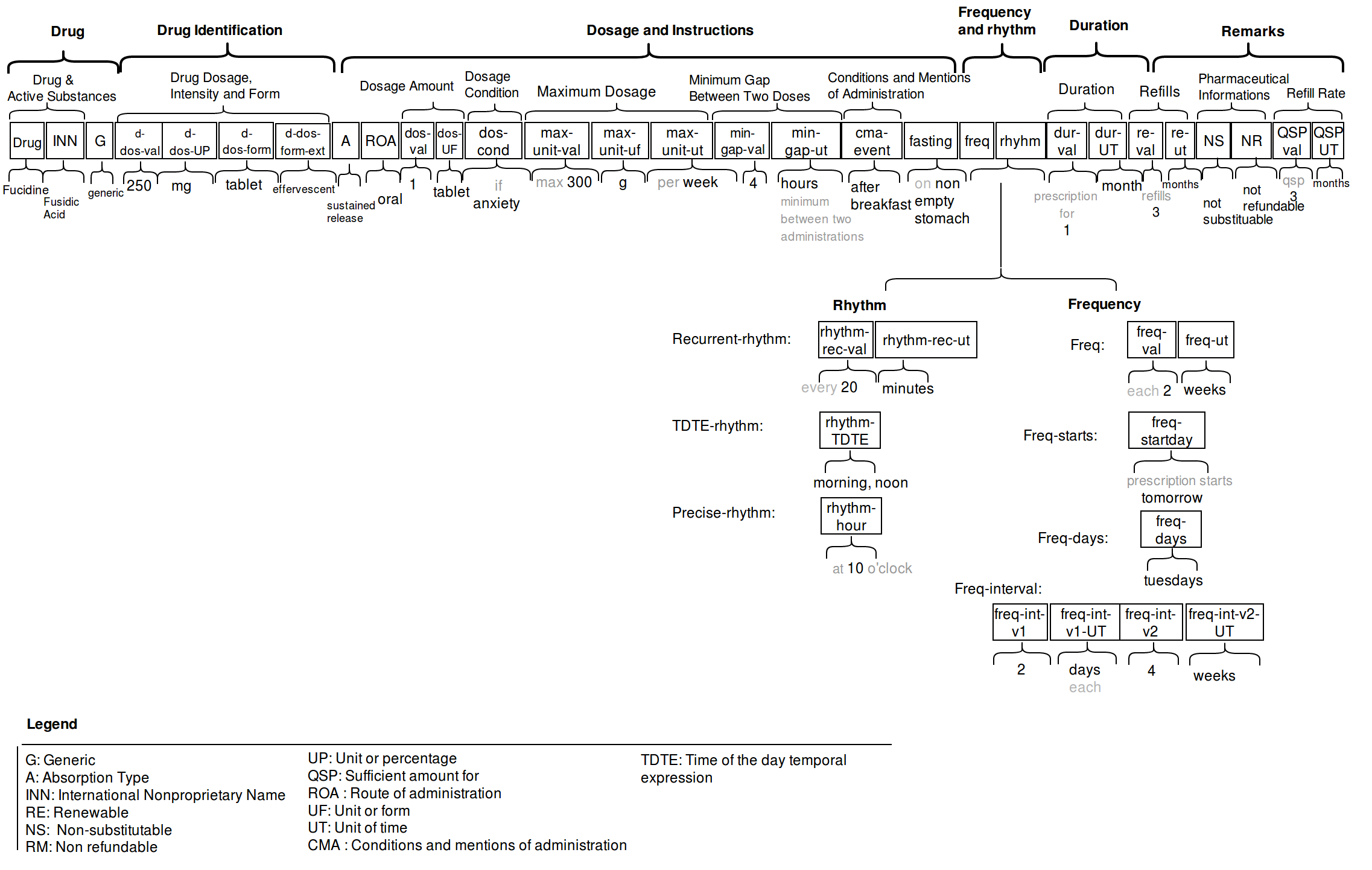}
    \caption{Slot-label definitions of the prescription domain with example values}
    \label{fig:domain}
\end{figure}

Figure~\ref{fig:domain} illustrates slot-labels with an example for each semantic category. This taxonomy includes 39 slot-labels and 269 slot-values describing the medical drug prescription domain. For instance, in the example  Figure~\ref{fig:dialogueSteps}, for the utterance ``Ofloxacine 200 mg\dots'', the unit of the drugs dosage which is mg is associated to ``{\tt d-dos-up}'' as slot-label while ``200'' is associated to the drug dosage value slot ``{\tt d-dos-val}'' 

\subsection{Data acquisition and data augmentation}\label{subsec:dataSped}

As stated in the introduction and related work sections, there is no available dataset of drug prescriptions in French. However, there is a large number of prescription textbooks for medical studies. Thus, to get access to realistic prescriptions, our strategy was to automatically extract drug prescriptions from the ``\emph{Le Guide des Premières Ordonnances}'' \cite{GuideOrdonnance2008} textbook that has been bought by the authors. The book has been digitized to a pdf version from which prescriptions were automatically extracted. 

832 textual prescriptions were extracted and automatically pre-annotated with the semantics defined in Section~\ref{subsec:semantics} using a set of regular expressions. All these annotations were then manually checked so that mistakes due to noise in the pdf extraction and to errors in the automatic annotation were corrected. The semantic definition contains 39 slot-labels, but of course drug prescriptions generally contains a much smaller amount of concepts. Furthermore, drug prescriptions contain some concepts common to most of the prescriptions (drug name, dosage, etc.) with extra specific information when needed. Table~\ref{tab:frequentSlots} reports the frequency of slots resulting from the processing of the textbook.

\begin{table}[htb]
    \centering
    \input{./freqSlotsRare.tex}
    \caption{Excerpt of frequency of some slots in textbook}
    \label{tab:frequentSlots}
\end{table}

Table~\ref{tab:frequentSlots} shows that the distribution of class labels is very imbalanced in practice. This situation is well known in machine learning and is called class imbalance. If not taken into account, the models learned from imbalanced data will be biased against majority classes. Several techniques exist to deal with this situation, from loss weighting to data augmentation using DNN \cite{kobayashi2018contextual}. However, loss weighting does not solve the data paucity and stochastic data augmentation needs an initial set of data which is still too important in our case. Hence, to solve the paucity of data and the class imbalance at the same time, we set up a simple generation technique. 

We defined a feature-based context-free grammar for the drug prescription domain. Top-level rules of the grammar include different parts of the prescription described in \ref{subsec:semantics}, whereas the terminals of the grammar are triplets of keywords, slot-label, and slot-value. For the generation, we created 10 high-level, 96 intermediate level and 72 terminal rules. Terminal rules are only 72 because, they were mostly composed of keywords replaced by values coming from the medical knowledge base according to the characteristics of the drug chosen randomly. For example, if the drug chosen for the prescription generation is paracetamol and tablets, the rest of the values are chosen in coherence with that form (ex. oral route, etc.).
The generator produces prescriptions containing underrepresented slots by dynamically converting slot-labels to top-level expansion rules of the grammar. Using half of the prescriptions acquired from \cite{GuideOrdonnance2008} as our initial training data, the slot-label distribution is computed to identify candidate slots which are underrepresented in the training set. Until a balanced distribution of slots is reached, random candidates are iteratively chosen to be produced as prescriptions. Finally, drug information is extracted from the French public drug database\footnote{\tt{http://base-donnees-publique.medicaments.gouv.fr}} in order to fill the triplets before adding the full prescription to the training data. At the end of the process, 3034 examples were generated using the grammar. For more details about the generation process, the reader is refereed to \cite{kocabiyikoglu2019towards}.

\section{Dialogue System}\label{sec:dialogue}

In this section, we define how our dialogue system fits in the global ecosystem of health dialogue systems and detail the modeling and the development process of different modules. 

\subsection{Overall architecture and positioning of the system into the ecosystem}

\begin{figure}[!hbt]
    \centering
    \includegraphics[width=\textwidth]{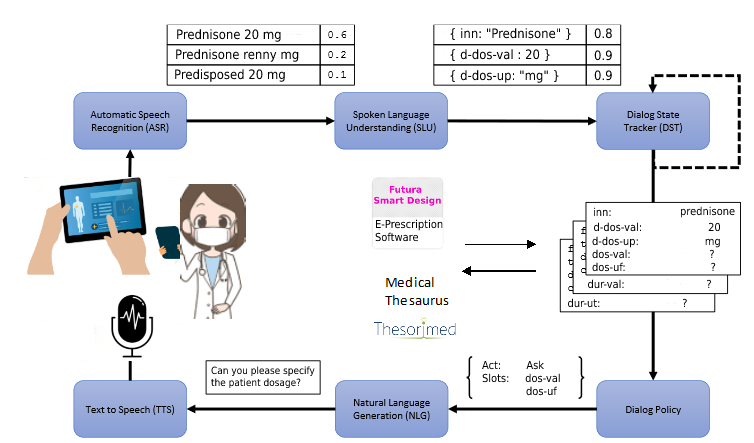}
    \caption{The pipeline of the modular dialogue system (adapted from \cite{williams2016dialog})}
    \label{fig:systemModulaire}
\end{figure}

Our goal-oriented dialogue system is destined for medical practitioners. Figure~\ref{fig:systemModulaire} illustrates the anatomy of a goal-oriented modular spoken dialogue system. The processing scheme (pipeline) starts with the automatic speech recognition (ASR) of the user's speech. The ASR module transforms the speech signal of an utterance into a textual transcription. Then, the NLU module extracts the semantic slots from this transcription as well as the speaker's intention. This meaning representation and the intention extracted from the transcription are passed to the dialogue state tracking (DST) module. This module is also referred to as the dialogue manager and allows the system to determine the dialogue state~\citep{williams2016dialog}. Given the state of the conversation, the system responds to the user with an answer determined by the dialogue policy which consists in choosing the action of the system adequate to the dialogue state. Once the system response determined, the system action is transformed into natural language by the natural language generation (NLG) module and then given to a text-to-speech (TTS) module which is returned to the user as a spoken utterance.

\begin{figure}[!htb]
    \centering
    \includegraphics[width=\textwidth]{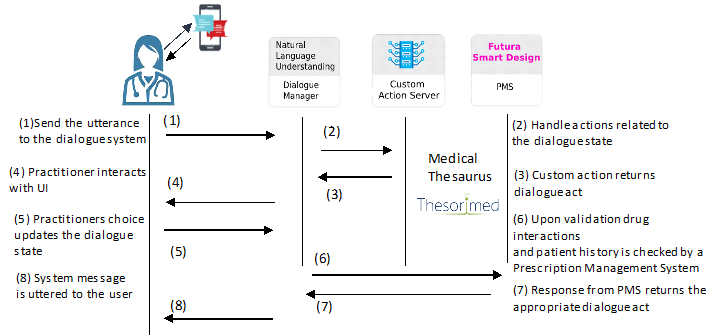}
    \caption{Interaction between the prescriber, dialogue server and external services
    }
    \label{fig:servers}
\end{figure}

The mobile phone is carried by the prescriber who can use it at any time. The prescriber alternates between voice and touch interaction depending on the status of the dialogue. This interaction is sent to a central server within the hospital that handles all the processing of the dialogue. This server is connected to a drug database (here, Thesorimed) for drug disambiguation and the e-prescription system (here, Futura Smart Design) for interaction with the EHR of the patient. Figure~\ref{fig:servers} gives an overview of the different components that involved in the prescription process. The drug database is Thesorimed\textregistered \vphantom{}, a large database regularly updated by the French National Health Insurance (Ameli). Two levels of verification of the prescription exist to ensure a secure information validation process. To do so, the processing pipeline involves different levels of verification using several medical knowledge bases. 

At first \textbf{(1)}, when a prescription intent is inferred from an utterance, slots are extracted \textbf{(2)} to be associated with common dispensation codes (UCD) of drugs (3). If no drugs can be associated to the utterance, the system suggests restarting the prescription process. If there is more than one drug candidate, the user is provided with a list of drugs to choose. These external queries to knowledge bases or other APIs are realized by the custom action server which has the objective to communicate with external services. The process then continues until all missing information is inferred (\textbf{4} and \textbf{5}). In the second stage, when the necessary prescription information is complete, this structured data is sent to the e-prescription software \textbf{(6)}. For a given patient file, e-prescription software handles the validation process of the prescription and give information about drug interactions, patient allergies, and so on (\textbf{7} and \textbf{8}). At the end, the practitioner validates the prescription or cancels it.

\subsection{Dialogue modeling}\label{sec:dialogue_modeling}

Since there is no existing practice in dialogue drug prescriptions, we started the dialogue modeling by analyzing the different steps of certified e-prescription software. 
In this software, the first step is to identify the medical drug, then identify a valid posology including the dosage, recommendations and frequency of dosing and finally, the duration of the posology. This allowed us to partition a dialogue session in various dialogue acts following an iterative process with an expert. 

\begin{table}[!hbt]
    \centering
    {\footnotesize
    \input{./dialActs.tex}
    }
    \caption{Summary of the dialog acts defined in the system (S=System, U=User)}
    \label{tab:dialogueActs}
\end{table}

Table~\ref{tab:dialogueActs} presents an overview of the dialogue acts we have defined. Thirteen different dialogue acts were defined regarding the prescription, correction, validation and removal of information. The task indicated in Table~\ref{tab:dialogueActs} shows the intention of the user, in our case, it could correspond to the task of prescribing medicine, the intention of restarting the session, etc. These dialogue acts were inspired from \cite{traum1992conversation} and was reviewed with medical experts in the iterative approach described in \ref{sec:method}.

To design the dialogue structure we designed seven scenarios with a domain expert. The seven scenario enabled to validate the structure of the dialogue acts and slots. Table~\ref{tab:scenarios} shows two examples of scenarios. Scenario~\textbf{A} represents a situation where the prescriber initiates the dialogue using a specialty name of a drug and its dosage. Once the information is extracted, the action \texttt{action\_check\_drug} \textbf{(1)} calls the disambiguated search of the drug candidates in a database. Then, the action \texttt{prescription\_form} \textbf{(2)} asks for the missing information until the prescription can be validated. scenario~\textbf{B} starts in the same way but does not require an explicit request for the drug thanks to disambiguation. However, once the prescription is submitted for validation, the user asks to delete an information which leads to the action \texttt{negate} \textbf{(3)}. The system then asks again for this mandatory information in the prescription. 

\begin{table}[!hbt]
    \centering
    {\scriptsize
    \input{./scenarios.tex}
    }
    \caption{Two dialogue scenarios}
    \scriptsize{* indicates information entered by the user.}
    \label{tab:scenarios}
\end{table}

\subsection{Spoken Language Understanding}\label{subsec:NLU}


While early slot-filling systems were rule-based \cite{tur:2011}, modern methods are data-driven. Conditional random fields \cite{jeong2008triangular} have been superseded by deep neural networks, including basic RNNs \cite{mesnilUsing2015}, Bi-directional LSTM RNN encoder-decoders \cite{Bapna2017}, Attention-based RNNs \cite{liu:2016}, Attention based CNNs \cite{huangImproving2017} and transformer based pre-trained models~\cite{lee2020biobert}. 
While intent detection has traditionally been seen as a separate task from slot-filling \cite{tranHierarchical2017}, since both tasks are highly correlated, some approaches perform slot-filling (sequence labeling) and intent detection (sequence classification) simultaneously. Such work includes Tri-CRF \cite{jeong2008triangular}, which extends the linear sequence labeling CRF with a node to represent the dialogue act, and Att-RNN \cite{Liu2016}, which extends the slot-filling encoder-decoder RNN with an extra intent decoder.

Given the low amount of data available to train NLU models, we considered four different approaches to address the slot-filling task: CRF, Tri-CRF, Att-RNN and Flaubert. Tri-CRF~\cite{jeong2008triangular,jeongMulti-domain2009} is an extension of a linear chain conditional random fields (CRF). Linear CRFs model the conditional probability distribution of the output label sequence, given the input sequences (sentences): each observed word $x_t$ in a sequence is conditionally dependent on its corresponding \textit{unobserved} label $y_t$. The Attention RNN (Att-RNN) model \cite{Liu2016} is a recurrent encoder-decoder architecture for simultaneous intent detection and slot labeling. The encoder is a bi-directional LSTM RNN which takes as input the sequence of words in an utterance, with one word $x_t$ input at each time step. Output $t$ each time step is the hidden state $h_t$ of the bidirectional RNN. Finally, Flaubert is a pre-trained transformer based contextual word representations for French that can be fine-tuned on supervised tasks~\cite{le2019flaubert}.

\subsection{Drug Retrieval}

    Identifying the drug of the prescription is the first important task of the dialogue. In classical e-prescription systems, practitioners choose drugs from an exhaustive list of drugs by querying keywords and dosages. Each drug of the list is associated with a unique commercial drug identifier.  However, in a spoken medical prescription system on mobile phones, this solution would be inadequate for a small screen and too slow. Hence, we decided to infer the right drug from the utterances. To do this, we used the extracted semantics from the utterance to match a national drug identification number (UCD code in France) and its international non-proprietary name (INN). 

Matching semantics to unique identifier names automatically is not an easy task because drugs are not always fully described and the given information varies from drug to drug. For example, for the drug ``\textit{Celluvisc\textsuperscript{\textregistered}}'', the only semantic category we need is the commercial brand name (drug) because it exists in only one dosage. On the other hand, for the prescription  ``\textit{Doliprane\textsuperscript{\textregistered} 500 milligrams tablets}'', even though we have more semantic information, the system cannot associate a unique drug id because there are two possibilities: regular tablets or effervescent tablets. This ambiguity in drug description can be dealt with by a pharmacist who will infer implicitly the drug described in the prescription. But, for an e-prescription software, the drug must be explicitly selected. 

\begin{figure}[!htb]
    \centering
    \includegraphics[width=\textwidth]{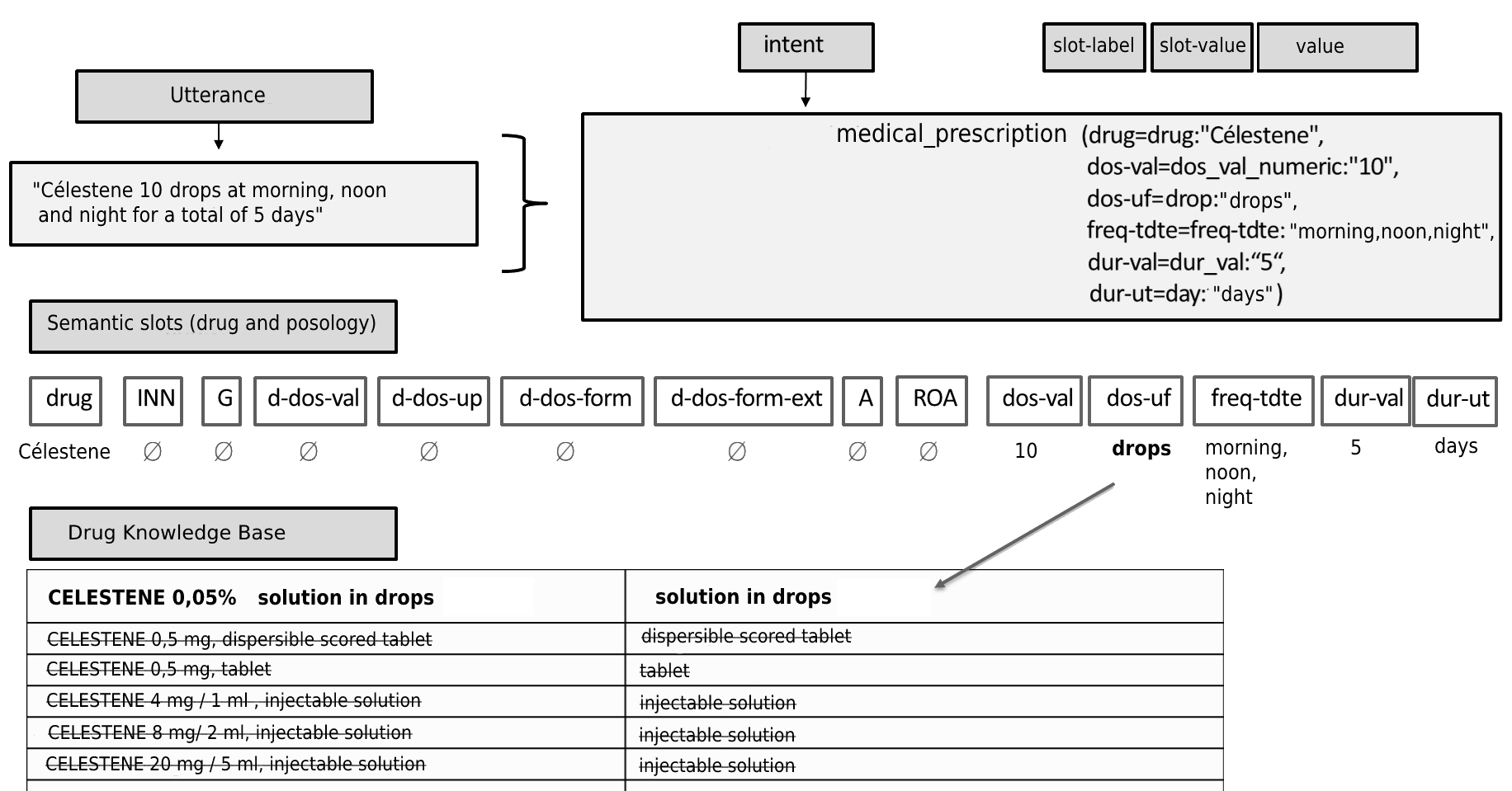}
    \caption{Example of the disambiguation process of a prescription.}
    \label{fig:desambigEx}
\end{figure}

In our case, for an oral utterance, we propose a disambiguation process using information from slots already acquired in the dialogue session. These slots are used to compute a similarity with the entries of Thesorimed{\textsuperscript{\textcopyright}}. Our API allows slots to be associated with a unique UCD (\textit{Unité Commune de Dispensation}) code. The disambiguation process uses regular expressions in an iterative way starting by a query with the name of the drug or the INN. If the result gives more than one answer then the query is extended with the dose, then with the dosage form, then with the route of administration, etc. according to the available attributes. If no answer is found, the user is prompted to start again. If only one answer is found, the system moves on to the next step of the dialog. If multiple answers are returned then the user is prompted to select an item from the list of possible drugs. 

Figure~\ref{fig:desambigEx} shows an example of this disambiguation process. In the example, the practitioner did not give any explicit semantic information about the drug, however, from the posology ``10 drops at morning, noon and night'', it is possible to infer that form of the drug could be a drinkable solution in drops and thus allows the disambiguation algorithm to filter out other possibilities. Given this utterance, the dialogue can pursue and associate the drug to a unique UCD code and later on confirm on the screen explicitly the full name of the drug when all of the prescription information is gathered.

\subsection{Dialogue Policy}\label{sec:dialogue_policy}

\begin{figure}[htb]
    \centering
    \includegraphics[width=\textwidth]{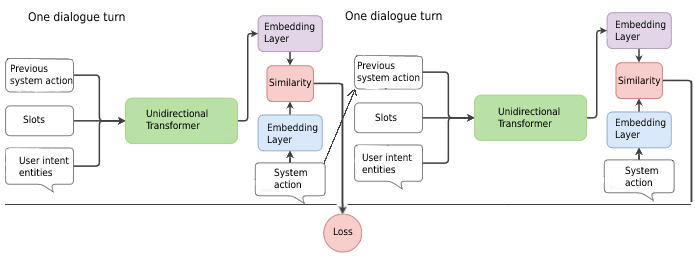}
    \caption{Schematic representation of transformer embedding dialogue policy adapted from \cite{vlasov2019dialogue}}
    \label{fig:transform}
\end{figure}

The dialog policy consists of choosing the best action to take based on the current utterance, the history and the current dialogue state. In this work, this policy is based on~\cite{vlasov2019dialogue} which uses a \emph{Transformers}~\cite{vaswani2017attention} architecture. in this approach, rather than considering the entire history and state of the dialogue, the \emph{self-attention} mechanism focuses the weights only on information relevant to the current state of the dialog for making a decision. This allows the system to better handle out-of-domain statements and non-cooperative scenarios. Figure~\ref{fig:transform} represents the processing of two cycles using this architecture. The current slots, utterance and the previous actions are given to a transformer model that computes the embeddings of the possible actions. Note that in this self-attention mechanism, the attention is on the sequences of dialogue turns and not on the lexical terms.  During the inference phase, the similarity score of the current state of the dialogue is computed and compared with all the actions of the system~\cite{vlasov2019dialogue}.

The output of the transformer and the system actions $y_{action}^{+}$ are embedded into a single semantic vector space $h^{+}_{action} \in \mathcal{R}^{20}$ in order to obtain the positive similarity $S^+ = h^{T}_{dialogue} h^{+}_{action}$. Similarity of negative actions are obtained in the same way $S^{-} = h^{T}_{dialogue}h_{action}^{-}$ and a negative action $y^-_{action}$. The sum is taken over the set of negative samples $\Omega^-$. For one dialogue, the loss function is:
\begin{equation}
    L_{dialogue} = - \langle S^{+} - \log{(e^{S^+} + \sum_{\Omega^-} e^{S^{-}})}  \rangle
\end{equation}
where the sum is taken over the set of negative samples $\Omega^-$  and the average $\langle .\rangle$ is taken over time steps inside one dialogue.
During training, $L_{dialogue}$ permits to maximize the positive similarities and minimize the negative similarities. The global loss is obtained as an average of all loss functions from all dialogues. For more information about the dialogue policy, the reader is referred to \cite{vlasov2019dialogue}.

\subsection{Interaction with the prescription system}

\begin{figure}[bht]
    \centering
    \includegraphics[width=0.7\textwidth]{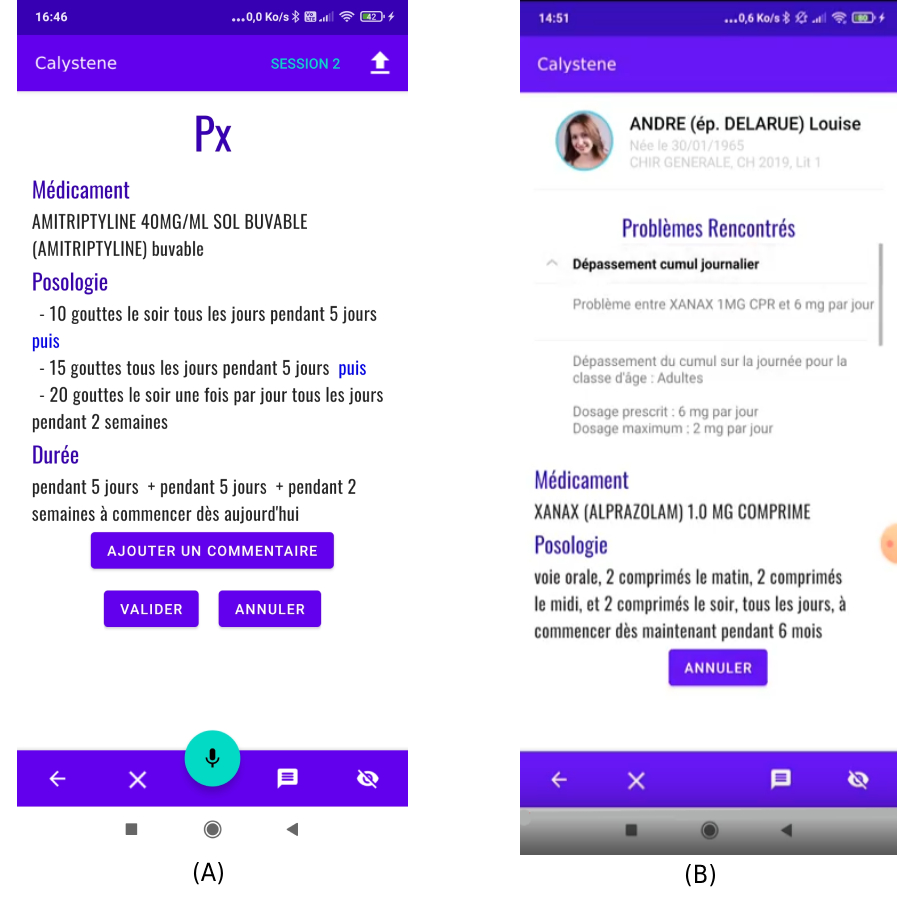}
    \caption{Two screen captures taken from the actual mobile application. (A) Validation of the meaning of the captured prescription by presenting to the prescriber. (B) validation phase of the prescription WRT the EHR of the patient, here the e-prescription system reports the quantity per day is too high.}
    \label{fig:interfaceMobile}
\end{figure}

For the dialogue flow and the verification process, we employed several strategies. The first is to ensure the name of the drug. The dialogue system tries to associate the semantics with a drug or proposes a list of drugs when there are several candidates. If the system is unable to identify a drug, it proposes to restart the prescription session (\textit{request\_restart}). When the system identifies the drug and the necessary slots (dose, duration, frequency, etc.), the system displays the prescription for the validation. Before the validation, the identified slots are not visualized. Figure~\ref{fig:interfaceMobile} shows two screen captures of two medical prescription examples in French from the deployed dialogue systems.
The recordings are sent through the dialogue system using the push-to-talk button that  can be seen at the bottom of Fig.~\ref{fig:interfaceMobile}~\textbf{(A)}.

Fig.~\ref{fig:interfaceMobile}~\textbf{(A)} shows an example of a prescription recorded within the data collection protocol. At this step, all the mandatory information has been captured and the prescription is summarized to the prescriber. The prescriber can choose to validate it as is, cancel it or add comments (e.g. in a big glass of water). The dialogue steps follow the processing scheme we described in  section~\ref{sec:method}. 
Fig.~\ref{fig:interfaceMobile}~\textbf{(B)} shows the feedback of an e-prescription software with an example of a prescription (Alprazolam). We can notice that, with this patient (the name and image are fake), the validation of the prescription triggers an alert in the e-prescription software. In this case, the given dosage for the patient has a daily accumulation that exceeds the usual recommendations. The prescriber is invited to either change their prescription or add a comment justifying their choice (not shown here).

\section{Experiment and corpus collection}\label{sec:exp_corpus}

To evaluate the dialogue system we ran an experiment between January 2021 and October 2021 (10 months of experimentation). This experiment had two main objectives. 
\begin{itemize}
    \item Evaluate our initial dialogue system prototype.
    \item Collect speech data for training models and sharing an annotated dataset to the community.
\end{itemize} 

This section describes how the prototype models were trained, the experiment protocol and the collected data.

\subsection{Models Training}\label{sec:training_data}

\textbf{NLU Models.\\}

The NLU models were trained on a dataset composed using the sources and method described in  Section~\ref{subsec:dataSped}. Table~\ref{tab:recapCorpusStats} shows the size of the train and development sets. It is composed of three sources: Textbook data which is coming from prescriptions extracted from \cite{GuideOrdonnance2008}, artificial prescriptions generated using the context-free grammar based generator (cf Sec.~\ref{subsec:dataSped}) and ESLO2 corpus\cite{serpollet2007large} which is composed of spoken French casual sentences.

\begin{table}[tbh]
\begin{small}

    \centering
\begin{tabular}{|l|r|r|} 
\hline
\multicolumn{1}{|c|}{\textbf{Source}} & \textbf{~~~train~~~} & \multicolumn{1}{|c|}{\textbf{dev}}   \\ 
 & \multicolumn{1}{|c|}{(\#utterances)} & \multicolumn{1}{|c|}{(\#utterances)}   \\ 
\hline
Textbook                              & 732            & 100            \\ 
\hline
Artificial~data~~~                       & 33028          & 300            \\ 
\hline
ESLO2                                 & 1416           & 100            \\ 
\hline
\hline
Total                                 & 35176          & 500            \\
\hline
\end{tabular}
    \caption{Distribution of the utterances of the drug prescription corpus according to source for NLU training}
    \label{tab:recapCorpusStats}
    \end{small}
\end{table}

Table~\ref{tab:corpusNlUStats} presents an overview of this training data with respect to the intents. It is composed mostly of synthetic sentences generated by our grammar. 832 examples out of 8833 regarding drug prescriptions comes from the textbook. We have also added an intent ``None'' to detect utterances out of the dialogue goal (e.g. the prescriber taking about another subject while entering a prescription). This none intent is composed of the ESLO2 corpus which contains only casual speech. All of the other intent examples were generated automatically. 

\begin{table}[bht]
    \centering
    \input{./nluIntentStats.tex}
    \caption{Distribution of the intent classes in the corpus of Table~\ref{tab:recapCorpusStats}}
    \label{tab:corpusNlUStats}
\end{table}

For the Tri-CRF model, to reduce training time, we pruned low-probability intents ($<$ 0.1\%) and initialized the weights using the pseudo-likelihood (for 30 training iterations). Training proceeded for 200 iterations. For Att-RNN, the input words were first passed to a 128-unit embedding layer. The bi-directional LSTM encoder and decoder were each a single layer of 128 units. Training was performed using stochastic gradient descent (SGD) with a batch size of 16, using gradient clipping at a norm of 5.0, dropout with a keep-probability of 0.5 and training was allowed to continue for 30,000 training steps. We selected the trained model with the highest F1 score on the slot labeling task on the validation set. 
The Rasa configuration, `spacy\_sklearn', uses a linear chain CRF to classify slot-labels 
. Separately, the model uses a linear SVM based on pre-trained word-embeddings to classify intents. For the transformer based contextualized word embedding, we used the Flaubert model (``\textit{flaubert-base-cased}'') which was trained on general purpose French corpora \citep{le2019flaubert}. This model was fine-tuned on the medical prescription training set for 3 epochs using the `huggingface' library.
~\\

\textbf{Dialogue Model.\\}

For the training of the dialogue system, all of the actions (user's actions and system actions) were encoded as a bag of words~\citep{vlasov2019dialogue}. For example, the action of the drug checking system was encoded as \textit{`$action_check\_drug' = \{action,check,drug\}$}. In our case, 31 actions were considered. Regarding the user's actions, in addition to the slots defined in the Section \ref{subsec:semantics}, we encode other slots internal to the system such as the UCD code, the validation status of the prescription, etc. The set of actions, slots and intentions led to $147$ parameters.
The architecture of the transformer is the one presented Section~\ref{sec:dialogue_policy} with the configuration used by \cite{vlasov2019dialogue}. The unit size of the transformer is 128 with 1 layer and 4 attention heads with an embedding dimension of size 20. Concerning regularization, we do not apply a dropout for attention but a dropout of 0.1 is applied for embedding training.

For the learning of the dialogue policy, we used a complete artificial generation of data. Starting from seven scenarios (cf. Sec~\ref{sec:dialogue_modeling}), we generated 14,255 dialogue sessions by filling in automatically slots with medical information. However, since these are generated from only seven scenarios, the variation remains small and most scenarios are cooperative dialogues. We therefore limited the training to 20 epochs only with a batch size of 32 that was increased linearly over the epochs up to 64. Given the nature of the task, we encoded a history limited to 10 rounds during training. Convergence was reached at the 16th epoch. 

\subsection{Data Collection Protocol}\label{subsec:protocol}


To allow participants to perform the data collection in their own environment, we deployed a dedicated server that allowed remote participation and data retrieval. This strategy enabled to collect data in a much more ecological way than inviting participants in the lab in order to record interactions in a dedicated room with an experimenter and also enabled us to perform experiments without breaching the sanitary protocol during the COVID-19 pandemic.
Thanks to a fruitful collaboration with the University Hospital of Grenoble, we were able to involve physicians and pharmacists despite the pandemic. Non-expert participants were native French who were either members of the lab of within the social network of the co-authors. 

Our goal was to reach about 30 naive users and about 30 experts in drug prescriptions, including some physicians. For this, we established a simple protocol were participants went through the following steps.

\begin{enumerate}
    \item Registration on the form for requesting to participate in the experiment.
    \item Reception of the .apk (installation file of the mobile application on Android) and reading of the document explaining the installation and the course of the experiment.
    \item Reception of prescription examples (depending on the audience: 20 reading examples for naive users; 10 pictographs and 10 reading examples for medical experts).
    \item Agreement to the terms of use and completion of the experiment.
\end{enumerate}

\subsection{Data Preparation}

Our data collection protocol targeted two types of participants: experts medical practitioners (physicians, pharmacists, biomedical engineers, nurses, etc.) and non-experts. Since all the participants did not have same expertise on medical prescriptions, we did not provide the same stimuli for these two categories. 

\begin{figure}[htb]
    \centering
    \includegraphics[width=0.99\textwidth]{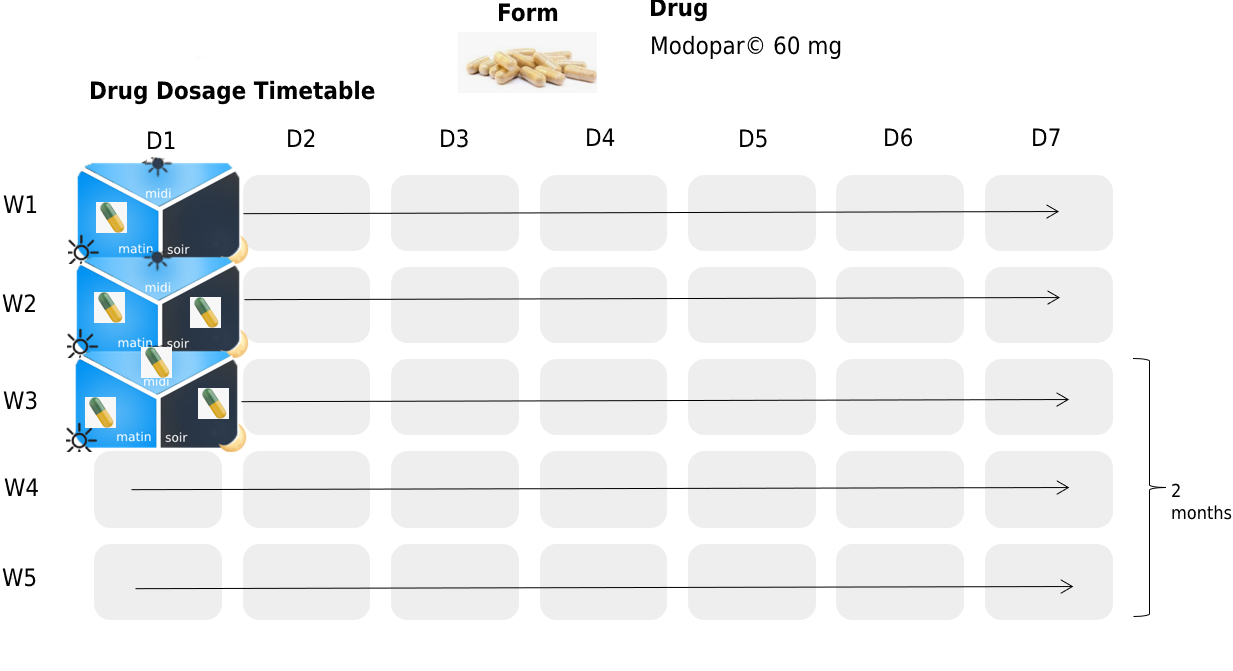}
    \caption{Example of a pictograph representation of a medical prescription}
    \label{fig:pictogramme}
\end{figure}

For non-expert users, we prepared a reading exercise that did not require any domain knowledge. However, for experts we wanted to be as close as possible to their own verbalization. Hence, we defined a method using iconic representations. Indeed, in order to prevent influencing the expert’s utterance, we provided representations of drug prescriptions in the form of diagrams that approximates drug taking timetables. These timetables are designed for patients, generally those who are taking multiple medications, to remind the dosage and times for each medication associated with some conditions and constraints. Figure~\ref{fig:pictogramme} shows an example of this representation. Such graphical representation allows limiting linguistic priors during the experiment.

In Figure~\ref{fig:pictogramme}, the drug (Modopar~\textcopyright) is explicitly given in written text. However, in order not to influence prescribers with the days of the week, the days are represented as (D1,D2,D3,D4,D5,D6 and D7). The dosage form of the drug is represented with an image, which in our example represents capsules but nothing prevents a participant to refer to it as pills or tablets. The names of the drugs, the conditions or the administration details are given in text form at the top of the screen for the prescriber to incorporate into their prescription. The dosage is indicated in the form of a calendar with boxes indicating the start time and their continuity in time. The dosage indicated in the example below denotes the progressive taking of one capsule of the drug in the morning for 1 week, then one capsule in the morning and evening for another week, then one capsule in the morning, noon and evening for 2 months.
Even though this representation allows prescribers to record prescriptions that are more natural, it has the disadvantage of taking more time than a reading exercise. That is why, after 10 pictographs, the experts are asked to read 10 new examples. The non-experts are directly provided with 20 textual stimuli to read. 

To prepare the stimuli, real examples of prescriptions were extracted from textbooks in medicine different from the one used for the training dataset such as \cite{schlienger2013100}, \cite{delcroix2020ordonnances}, \cite{ordonnances180}, \cite{ordonnancesParasitologie}, \cite{delcroix2020ordonnances} and discussed with our experts (two of which are co-authors of this paper). When prescriptions were not complete, we added duration, rhythm and frequency information with plausible values. Given the number of participants targeted, the material generated for the experiment represented approximately 300 examples of pictograph and 1300 textual drug prescriptions. Our preparation included ranking of the prescription according to their complexity (i.e., the longest and the ones with several dosage changes were last).

\subsection{Corpus Overview -- PxCorpus}

The result of the experiment, enabled us to collect a drug prescription dataset that we cleaned, annotated and distributed under the name of \textbf{PxCorpus}. Figure~\ref{tab:statsCollecte} gives an overview of the dataset. It includes 1981 recordings of 55 participants (38\% non-experts, 25\% physicians, 36\% medical practitioners) with the data manually transcribed and semantically annotated. In total, it contains $4h$ of speech recordings acquired from human-machine interactions using our dialogue system as a prototype.

\begin{table}[!htb]
    \centering
    \input{./CorpusOverview.tex}
    \caption{Overview of PxCorpus}
    \label{tab:statsCollecte}
\end{table}

\begin{figure}[!htb]
    \centering
    \includegraphics[width=\textwidth]{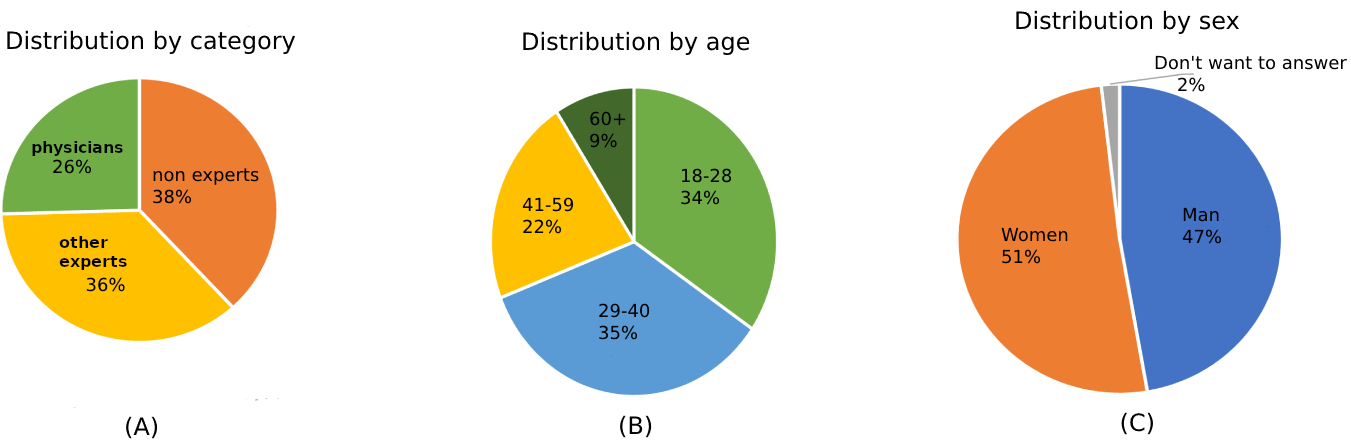}
    \caption{Distribution of some characteristics of the participants}
    \label{fig:repartitionParticipants}
\end{figure}

Figure~\ref{fig:repartitionParticipants} shows the distribution of characteristics of the participants. \textbf{(A)} shows that the collected dialogues are evenly distributed among the participant categories. The pie chart in \textbf{(B)} represents the distribution of the data in relation to age ranges which shows that 3 age ranges are fairly represented while the over 60+ year-old range represents only 10\% of the participants. Finally, \textbf{(C)} shows that the gender representation (M/F) is well balanced. 

All the recording were transcribed (speech transcription) and semantically labeled (slot and intent) by two paid native speakers supervised by the co-authors during the summer 2021. The labeling process included 5 types of intents and 40 semantic labels that characterize drug prescriptions. See \cite{kocabiyikoglu2022} for more detail about the annotation process.  

The entire PxCorpus is made publicly available through a Attribution 4.0 International (CC BY-4.0) license at the following url: \url{https://doi.org/10.5281/zenodo.6524162}. It is distributed in an aligned format ready for developing and evaluating Spoken Language Understanding (SLU) systems (\textit{conll} format). This corpus is distributed in two parts, PxSLU for spoken language understanding presented in \cite{kocabiyikoglu2022} and PxDialogue. Speech recordings and semantic annotations are coming from the same source, however in PxDialogue the recordings are kept in the dialogic context which can allow to evaluate dialogue policies or to extract dialogue metrics. 

Table~\ref{tab:resumePxSLU} summarizes the annotations of PxSLU corpus distributed for the spoken language understanding task. 14068 instances of slot-labels and 1981 instances of intents were labeled. Only 5 slots out of the 40 slot labels had fewer than 12 instances. All other slots had from 5 to 1831 instances. For more detailed information about the intents and slot distributions, the readers can refer to \cite{kocabiyikoglu2022}.

\begin{table}[!htb]
    \centering
    \input{./pxSLUtab.tex}
    \caption{Summary of the annotations of PxSLU corpus}
    \label{tab:resumePxSLU}
\end{table}

Table~\ref{tab:resumePxDialogue} gives an overview of the PxDialogue corpus. PxDialogue corpus is composed of same utterances as the PxSLU corpus, with the difference that the dialog context (dialogue turns) are kept. The semantic and intent alignment are kept the same with PxSLU with an additional annotation for dialogues. We manually annotated each line of dialogue indicating when dialogue fails and excluded free-form comments that practitioners can add as remarks to prescriptions. Hence, this corpus can be used both for evaluation and training.  

\begin{table}[!htb]
    \centering
    \input{./pxDialTab.tex}
    \caption{Overview of the PxDialogue Corpus}
    \label{tab:resumePxDialogue}
\end{table}

\begin{figure}
    \centering
    \includegraphics[width=0.7\textwidth]{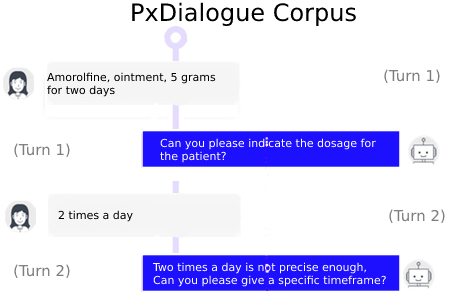}
    \caption{A dialogue sample illustrating dialogue turns}
    \label{fig:pxDialTurns}
\end{figure}

Dialogue turn given in the Table \ref{tab:resumePxDialogue} describes a consecutive turn of the user and the system. For example, the prescriber initiating a prescription and the system response which shows all of the drugs matching the drug description constitute a dialogue turn. As shown in the Figure \ref{fig:pxDialTurns}, a dialogue turn (turn x) is composed of a user action and a consecutive system action. In our case, user and system actions can be button clicks, push-to-talk recordings, TTS from the system, etc.)

\section{Results}\label{sec:results}

The experiment using our mobile prototype system allowed us to evaluate it in wild condition. In this section, we summarize the global performance of the system as well as of its different components. In particular we compare different NLU systems, extract metrics from dialogues that gave us an insight about the use of a mobile application for spoken medical prescription recordings via dialogue.

\subsection{NLU evaluation}

Table ~\ref{tab:resltsNLU} shows the results of the NLU systems trained on the corpus of section~\ref{sec:training_data} but evaluated on PxSLU corpus for both intents and slots. For intent, accuracy was used whereas the precision, recall and f-measure including micro, macro measures were chosen for slots. The macro average measures are very important since it considers all \textit{slots} of equal importance whatever their frequency. Indeed, according to the nature of prescriptions, most of the slots are optional and hence occur far less frequently than the mandatory ones.  

The results show that the performance of the model \textit{Flaubert} based on a pre-trained Transformer model gives the best results both for micro and macro level measures. All other models have comparable performance. We can see the same behavior for the intent accuracy. The lower macro average performances means that not all slots are recognized equally. However, from the macro average perspective, the Flaubert model is by far the more robust since it performs well even for slots which are rare. 

\begin{table}[!hbt]
    \centering
    {
    \input{./NLUResults.tex}
    }
    \caption{NLU models performance on the PxSLU Corpus}
    \label{tab:resltsNLU}
\end{table}

Apart from providing a real test-bed to evaluate NLU model, we wanted to check if the corpus could be used for fine-tuning the Flaubert model. For this purpose, we performed a K-Fold (K=5) cross validation with the pre-trained model ``\textit{flaubert-base-cased}''. In the cross-validation process, the dataset was iteratively split into $k$ roughly equal parts. In each iteration, each of the $k$ part was used as a test set and the rest was used for training. In each run, \textit{fine-tuning} was performed for three epochs. 

\begin{table}[!hbt]
    \centering
    {
    \input{./crossVal.tex}
    }
    \caption{K-fold cross validation result of the Flaubert model on PxSLU Corpus (SD=Standard Deviation)}
    \label{tab:crossVal}
\end{table}

Table ~\ref{tab:crossVal} shows the results of this experiment. We can see that a model trained on PxSLU obtain comparable results with those given in Table~\ref{tab:resltsNLU}. This shows that PxSLU can be used for fine-tuning and evaluation to lead to similar performance than models trained on larger artificial datasets (books as well as artificial data). 
Moreover, the results of the different k-folds vary more when we look at the macro average, which is confirmed by the standard deviation which is also higher. This shows that the choice of data impacts the macro performance and thus the coverage of the \textit{slots}. A finer data partitioning could thus lead to even greater performance. 

\subsection{Dialogue Metrics}

To evaluate the usability of the system, we extracted automatically statistics about the produced dialogues. Since the dialogue system was deployed as a mobile application, each button click, utterance, system response, etc. were saved as ``events'' along with timestamps. Table~\ref{tab:databaseDial} shows the information that was saved on \textit{sqlite} databases while participants were performing the experiment with a dialogue example. Here, a dialogue turn consists of a user utterance (\textbf{U}) followed by the consecutive system response (\textbf{S}).

\begin{table}[!htb]
    \centering
    {\scriptsize
    \input{./dialogueSqlite.tex}
    }
    \caption{Example of a dialogue session saved in local sqlite database of participants (TS = Timestamp in seconds)}
    \label{tab:databaseDial}
\end{table}

Using metadata saved in the database for each participant, we extracted the metrics summarized in Table~\ref{tab:metricsDialogue}. 

\begin{table}[!htb]
    \centering
    \input{./metricsPxDial.tex}
    \caption{Overview of PxDialogue Corpus Dialogue Metrics}
    \label{tab:metricsDialogue}
\end{table}

The task success rate indicates that according to all audiences, the task of spoken drug prescriptions was successful using the dialogue system for more than 70\%. The highest task success rate belongs to the non-experts category. This is probably due to the experimental setup which consisted only of reading a well-formed prescription. However, for the medical experts, task success was related to other factors (understanding or not the pictographs, the priority of the information). For the physicians, the success rate of the task was slightly higher than for medical experts.

The average time to complete a prescription session was approximately one minute for physicians, 50 seconds for non-experts, and 36 seconds for medical experts. In relation to this time, it can be noted that physicians and non-experts tried to correct the wrongly recognized slots in the more than other experts. This finding is confirmed by the higher average number of dialogue turns for physicians and participants than for other experts.  

We also measured the number of events during the dialogue sessions. Since in our case the interaction does not only consist of voice interaction, the events encompass any action performed on the smartphone (displaying a prescription, choosing a medication from a list, clicking on the back button, etc.).  We can see that the number of events correlates with the average time and the number of turns per session. Concerning metrics on user frustration, we measured the number of system errors, occurrences of repeat requests, and \textit{restart} requests from the system. Fatal errors (\emph{crash} system) were encountered in approximately 4 records for both naive users and physicians. Similarly, the ratio of error to dialogue turns (e.g. the system misclassified the users' intent) was less than 10\% for all participants. 

More specifically for spoken drug prescription task, one of the indicators that can cause frustration for participants is the rate of drug association. Indeed, if the medication described by the prescriber is not recognized, the dialogue must restart. The drug association rate shows that the percentage of dialogues where the drug is correctly found is 87\% for physicians and 95\% for other experts. Non-experts participants had the lowest association rate (82\%) which can be explained by the difficulty of uttering some drugs name. For other experts,  mostly composed of pharmacists, this high rate may be explained by the clarity of pronunciation of complex drug names or the more detailed drug descriptions that allowed a higher association rate.

\subsection{Dialogue Metrics by Category}

In order to have more specific measures, we analyzed the results according to the age and gender of the participants. Table~\ref{tab:resultsAgeFunction} shows the results on the dialogues produced according to their age category. 

\begin{table}[!hbt]
    \centering
    \input{./PxDialStatsbyAge.tex}
    \caption{Dialogue metrics by age category}
    \label{tab:resultsAgeFunction}
\end{table}

The first observation we can make is that the task success rate increases with age which comes at the cost of an increase in the time spent for a dialogue session. This shows that older participants have made more attempts to get a valid prescriptions. This is shown as well by the increase of the number of turns and number of events with age. We discovered that the younger participants got frustrated more easily and abandoned dialogue sessions more quickly. In contrast to the task success rate, the drug association rate decreases with age. This may be related to the fact that 57\% of the 18-28 years old participants are in the category of medical experts who are best recognized.

\begin{table}[!hbt]
    \centering
    \input{./statsDialSexe.tex}
    \caption{Dialogue metrics by gender}
    \label{tab:resultsGenderCategory}
\end{table}

The Table \ref{tab:resultsGenderCategory} shows the metrics extracted from dialogues according to participants gender. The results show that there is no significant difference in performance. However, it can be noted that the elapsed time for men is slightly higher than for women (on average 52 seconds versus 43 seconds).

\begin{figure}[!ht]
    \centering
    \subfloat[\centering Non-experts]{{\includegraphics[width=6.5cm]{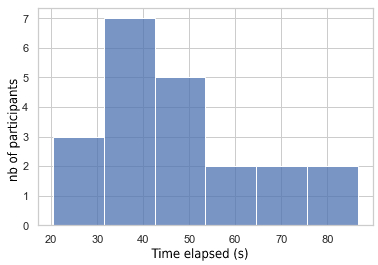} }}%
    \qquad
    \subfloat[\centering Physicians]{{\includegraphics[width=6.5cm]{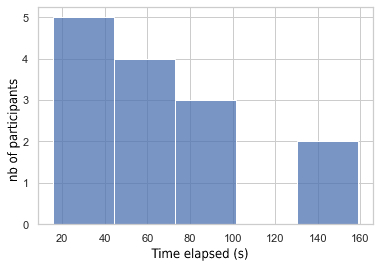}}}%
    \subfloat[\centering Other experts]{{\includegraphics[width=6.5cm]{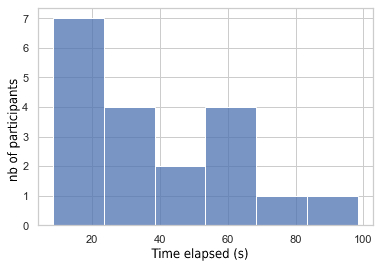} }}%
    \caption{Histograms of dialogue session times for each participant type}%
    \label{fig:dialogueTimeStats}%
\end{figure}

Since time is an important feature of the proposed system we included a more detailed analysis about the time elapsed for a dialogue turn in order to complete the task. The histograms of Figure~\ref{fig:dialogueTimeStats} show the sessions duration by participant category. Session duration is the difference between the timestamp of the last event and the first events timestamp for a given session. For physicians and non-experts, the time to complete spoken medical prescription entry is just under one minute except for a few participants who were slower than others. However, for the medical experts, we see that most participants completed sessions in less than 40 seconds.

\section{Discussion}\label{sec:discussion}

We presented a complete approach including NLU, dialogue modeling and evaluation for spoken medical prescription task. The results of the experiments described in the section~\ref{sec:results} show that drugs can be prescribed in less than a minute while ensuring compliance with a prescription with a success rate of over 72\%. 
Comparing such completion time with our spoken prescription system with actual e-prescribing software is difficult since there were only a few studies that have been reported in the literature. Furthermore, the targeted population, system and kind of prescription are very diverse. In \cite{devine2010electronic}, a study comparing hand-writing to e-prescription reported an average time of 69 seconds/prescription-event using the CPOE system in the examination room that is 25 seconds longer than handwriting. In \cite{10.1136/amiajnl-2012-001414}, the authors compared the time spent on different task of clinicians before and after the introduction of an electronic medication management system (eMMS). They found that the time spent on the prescription increased from 43.3 seconds in average to 65.3 seconds when using the eMMS system. Even though such comparison should be taken with care, our system exhibits an average time of 66.15 seconds for physician and 35.64 seconds for medical experts and the younger participants. These results support the hypothesis that with sufficient training time, it would be possible for prescribers to reach a few tens of seconds for prescriptions.

To build our system, we followed an iterative approach that allowed us to bootstrap a system, validate its modeling with experts, and obtain data for both realistic dialogue scenarios and training data on oral drug prescriptions. This system was interfaced with an e-Prescription software which allowed to obtain a real-time \textit{feed-back} mechanism through the dialogue. However, the evaluation focused on the input and validation of a prescription by the prescriber, but without any link with a specific patient. However, an e-Prescription software can provide more personalized information (e.g., allergies) or contraindications with current medications recorded in the patient's file. However, this integration also requires a data preparation phase including hospital units, patient profiles, pharmaceutical settings of drugs, etc.
A direct perspective of this work would be to evaluate the system in a full ecological context. This would confirm the evaluations of the system (which were realistically performed, but did not involve real prescription cases in a real ecological situation) while adding the patient dimension. This would allow us to check whether the additional interactions, especially when there are contraindications, still allow for a reasonable input time. Finally, in the long term, the integration with the e-Prescription software could measure whether the spoken interface has a positive effect on traceability and error reduction.


Concerning natural language understanding module, with the availability of pre-trained language models, especially trained on medical corpora, the performances of the different tasks have been significantly increased. To take into account the power of the transformer models, we have integrated \textit{Flaubert} only in this last version developed after the data collection. It would be interesting to see the impact of this model on the dialogue metrics, especially in an ecological evaluation. Moreover, our system was successfully bootstrapped using textbook and synthetic data. While this data generation method allows for an increase in the representation of the \textit{slots}, it remains rather unrepresentative of the reality of the field. Furthermore, the creation of rules for data generation is a time-consuming process. Now that we have realistic data, it would be interesting to test generative adversarial networks (GANs) for NLU or dialogue data generation. Recent studies \cite{golovneva2020generative} have shown that with a GAN approach, it is possible to generate NLU data even with limited resources.

For the dialogue system, the modular pipeline approach allows to put a control mechanism on the processes, brings traceability for errors. However, it generally implies a separate training of the modules, which makes them more sensitive to the errors which can have a cascade effect. Some approaches advocate merging certain modules. For example, \cite{desot2019slu,desot:hal-03727285} show that it is more advantageous to use an end-to-end SLU system where attributes and intent are extracted directly from a speech signal, than a pipeline system. An interesting research perspective could be to explore the following question: would it be possible to jointly optimize all modules of a dialogue system, while keeping separate models for each module? In the state of the art, we have presented some approaches that aim at optimizing modules through joint learning. For example, \cite{zhao2016towards} have shown that it is possible to jointly train the follow-up management (DST) and the dialogue policy. It would therefore be interesting to explore these approaches to both increase performance and keep the modularity that is very important for software maintenance.

\section{Conclusion}\label{sec:conclusion}

In this paper, we presented the full approach that enabled the creation of a medical drug prescription dialogue system to be used on a mobile phone. The aim of such system is to decrease time spend on e-prescription systems by providing a more natural interface and by enabling physicians to stay at the point of care. To the best of our knowledge, it is the first complete system of this kind. 

We approached the problem as a goal-oriented dialogue using slot-filling. However, since no such system exists, we had to both define the semantic space of the dialogue system and develop NLP models in a low resource setting. We thus made a first contribution to the field by elaborating an exhaustive taxonomy of the semantics of drug prescriptions expressed in natural language by co-construction with a domain expert.
Concerning the automatic understanding of prescription semantics, starting from a medical textbook of medical prescriptions, we proposed a method for data augmentation using an artificial data generation method that allowed us to increase the amount of data to balance underrepresented classes in the collected examples. This method allowed us to build a corpus balanced in terms of attribute distributions and large enough to use machine learning techniques. Thank to this method we obtained performances in-line with that of real data.
Finally, we presented an evaluation of the dialogue system in the wild with 55 participants. This evaluation showed that our system exhibits an average time of 66.15 seconds for physicians and 35.64 seconds for other experts and a task success rate of 76\% for physicians and 72\% for other experts. All these data was recorded during this evaluation. Such results can even be improved since further experiments showed that the NLU module can reach 94\% accuracy on the intent classification and 90\% F-measure for slot identification. In total, we collected about $4$ hours of sound recordings comprising over 2000 recordings. After transcription and semantic annotation, this PxCorpus of spoken drug prescription has been made available to the community to encourage reproducibility and support the development of NLU and dialogue models.

In this work, the focus was on medical drug prescriptions. However, the same methodology could be applied to other types of prescriptions (radiology, laboratory tests). It could therefore be interesting to learn NLU models on several types of prescriptions at the same time in the form of multitask learning task so that the knowledge acquired on one type of prescription can be transferred to another type. This type of transfer could also take place between different languages. For example, it would be interesting to explore bilingual approaches (English/French)\cite{rebholz2013entity} on corpora of medical reports or corpora of spoken or textual prescriptions. In this way, the small corpus acquired in this work and the vast MIMIC dataset in English could be leveraged to transfer knowledge from English to French.

\section{Acknowledgements}
This work was supported by a CIFRE grant number 2017/1798 from ANRT (National Association for Research and Technology) and was partially supported by MIAI@Grenoble-Alpes (ANR-19-P3IA-0003).



\bibliographystyle{unsrt}
 \bibliography{biblio}





\end{document}

%% file: challenges.tex
\begin{tabular}{|c|l|} 
\hline
\textbf{Task}                                                                                       & \multicolumn{1}{c|}{\textbf{Objectives}}                                                                                    \\ 
\hline
\begin{tabular}[c]{@{}c@{}}I2B2: Informatics for Integrating Biology\\ and the Bedside\end{tabular} & 2009 - Medication extraction                                                                                            \\ 
\hline
N2C2: National NLP Clinical Challenges                                                              & 2018- ADEs and medication extraction                                                                                    \\ 
\hline
SemEval                                                                                             & \begin{tabular}[c]{@{}l@{}}2013- Extraction of drug-drug interactions\\ 2014-15: NER and concept encoding\end{tabular}  \\ 
\hline
\begin{tabular}[c]{@{}c@{}}Medication and Adverse Drug Events\\ (MADE)\end{tabular}                 & 2019- Medication and ADE extraction                                                                                     \\
\hline
\end{tabular}

%% file: freqSlotsRare.tex
\begin{tabular}{|l|l|}
\hline
\multicolumn{1}{|c|}{\begin{tabular}[c]{@{}c@{}}Frequent slots\end{tabular}} & \multicolumn{1}{c|}{\begin{tabular}[c]{@{}c@{}}Rare slots\end{tabular}} \\ \hline
drug: 816                                                                           & inn : 17                                                                           \\ \hline
rhythm-perday: 427                                                                  & rhythm-rec-val: 5                                                                  \\ \hline
d-dos-form: 125                                                                     & d-dos-form-ext:5                                                                   \\ \hline
dur-val: 190                                                                        & re-val: 1                                                                          \\ \hline
\end{tabular}

%% file: dialActs.tex
\begin{tabular}{|c|c|c|l|} 
\hline
Dialogue Acts                                                                                   & S                     & U & \multicolumn{1}{c|}{Description}                  \\ 
\hline
\begin{tabular}[c]{@{}c@{}}\textit{inform(task=pres-drug,}\\\textit{ a=x,b=y,...)}\end{tabular} &                       & \checkmark & Dialogue started with the prescription intent   \\ 
\hline
\textit{request(a=restart)}                                                                     &                       & \checkmark & Explicit intent of session restart                \\ 
\hline
\textit{inform(a=drug\_not\_found)}                                                             & \checkmark                     &   & Inform that there is not match for the drug       \\ 
\hline
\textit{inform(a=drug,b=dos-val,...)}                                                           &                       & \checkmark & User informs the system of slots a=x,b=y,...      \\ 
\hline
\textit{ask(a=dos-val,b=dos-uf,...)}                                                            & \checkmark                     &   & System asks for slots a=x,b=y,...                 \\ 
\hline
\textit{affirm()}                                                                               &                       & \checkmark & User confirms the prescription                    \\ 
\hline
\textit{negate()}                                                                               &                       & \checkmark & User refuses the prescription                     \\ 
\hline
\textit{repeat()}                                                                               & \checkmark                     &   & System asks for user to repeat                    \\ 
\hline
\textit{reqalts(a=drug,,...)}                                                                   & \checkmark                     &   & System asks for alternatives for slots a=x,...    \\ 
\hline
\textit{request(a=comment)}                                                                     & \checkmark                     &   & System asks for a comment                         \\ 
\hline
\textit{inform(a=comment)}                                                                      &                       & \checkmark & User informs the system with a comment            \\ 
\hline
\textit{remove(a=x,b=y,....)}                                                                   & \multicolumn{1}{l|}{} & \checkmark & User asks to remove the values of slots a=x,   \\ 
\hline
\textit{replace(a=x, b=y,...)}                                                                  & \multicolumn{1}{l|}{} & \checkmark & User asks to replace the values of slots a=x,  \\
\hline
\end{tabular}

%% file: scenarios.tex
\begin{tabular}{|l|l|}
\hline
\multicolumn{1}{c}{\textbf{Scenario A}} & \multicolumn{1}{c}{\textbf{Scenario B}} \\ \hline
\begin{tabular}[c]{@{}l@{}}*medical\_prescription\{"drug","d-dos-val","d-dos-up"\}\\ \textbf{(1)} action\_check\_drug\\ * inform\{"drug", "d-dos-up", "d-dos-val", \\ "d-dos-form", "inn", "roa", "ucd\_code"\\ - slot\{"d-dos-form"\} - slot\{"d-dos-up"\} - slot\{"roa"\}\\ - slot\{"d-dos-val"\} - slot\{"drug"\} -  slot\{"inn"\}\\ - slot\{"ucd\_code\}\\ \textbf{(2)} prescription\_form\\ - slot\{"dos-uf"\} -slot\{"dur-val"\} -slot\{"dur-ut"\}\\ - slot\{"dos-val"\}\\ - slot\{"requested\_slot": null\}\\ - action\_confirm\_prescription\\ * inform\{"validate"\}\\ - slot\{"validate": "validate"\}\\ - action\_utter\_validation\end{tabular} & \begin{tabular}[c]{@{}l@{}}*medical\_prescription\{"drug","d-dos-val","d-dos-up"\}\\ - \textbf{(1)}action\_check\_drug\\ - slot\{"d-dos-form": "d-dos-form"\}- slot\{"d-dos-up"\}\\ - slot\{"d-dos-val"\}- slot\{"drug"\}- slot\{"inn"\}\\ - slot\{"roa"\}- slot\{"ucd\_code"\}\\ \textbf{(2)} prescription\_form\\ - slot\{"dos-uf"\}- slot\{"dur-val"\}- slot\{"dur-ut"\}\\ - slot\{"dos-val"\}- slot\{"requested\_slot": null\}\\ - action\_confirm\_prescription\\ * \textbf{(3)}negate \{"drug"\} OR negate\{"d-dos-form"\}(...)\\ - action\_negate\_delete\_item\\ \textbf{(2)} prescription\_form\\ - slot\{"dos-uf"\}- slot\{"dur-val"\}- slot\{"dur-ut"\}\\ - slot\{"dos-val"\}- slot\{"requested\_slot": null\}\\ - action\_confirm\_prescription (...)\end{tabular} \\ \hline
\end{tabular}

%% file: nluIntentStats.tex
\begin{tabular}{|l|r|} 
\hline
\multicolumn{1}{|c|}{\textbf{Intent class}} & \multicolumn{1}{|c|}{\textbf{train+dev}}  \\
 & (\#utterances)  \\ 

\hline
medical\_prescription                 & 8833                \\ 
\hline
request\_restart                      & 95                  \\ 
\hline
negate                                & 12608               \\ 
\hline
replace                               & 12624               \\ 
\hline
none                                  & 1516                \\
\hline
\hline
Total                                  & 35676                \\
\hline
\end{tabular}

%% file: CorpusOverview.tex
\begin{tabular}{|c|c|c|c|} 
\hline
\begin{tabular}[c]{@{}c@{}}\\\textbf{}\end{tabular} & \textbf{Sessions} & \textbf{Recordings} & \textbf{Time (m)}  \\ 
\hline
\textbf{Medical experts}                            & 258               & 434                 & 94.83              \\ 
\hline
\textbf{Doctors}                                    & 230               & 570                 & 105.21             \\ 
\hline
\textbf{Non experts}                                & 415               & 977                & 62.13              \\ 
\hline\hline
\textbf{Total }                                     & 903               & 1981                & 262.27             \\
\hline
\end{tabular}

%% file: pxSLUtab.tex
\begin{tabular}{|c|c|c|c|} 
\hline
\textbf{Utterances} & \textbf{Tokens} & \textbf{Slots} &  \textbf{Intents}  \\ 
\hline
1981               & 22440  & 14068                    & 1981                    \\
\hline
\end{tabular}

%% file: pxDialTab.tex
\begin{tabular}{|c|c|c|c|} 
\hline
\textbf{Dialogue Sessions} & \textbf{Dialogue Turns} & \textbf{User Turns} & \textbf{Recording time}  \\ 
\hline
959                        & 3675                    & 2067                & 262 minutes              \\
\hline
\end{tabular}

%% file: NLUResults.tex
\begin{tabular}{|c|c|c|c|c|c|c|c|} 
\hline
\multirow{2}{*}{\begin{tabular}[c]{@{}c@{}}\\\textbf{ Model}\end{tabular}} & \multirow{2}{*}{\begin{tabular}[c]{@{}c@{}}\textbf{Intent}\\\textbf{(acc)}\\\textbf{ }\end{tabular}} & \multicolumn{3}{c|}{\textbf{Micro Avg}}       & \multicolumn{3}{c|}{\textbf{Macro Avg}}        \\ 
\cline{3-8}
                                                                           &                                                                                                      & \textbf{P}    & \textbf{R}    & \textbf{F1}   & \textbf{P}    & \textbf{R}    & \textbf{F1}    \\ 
\hline
CRF                                                                        & 92\%                                                                                                 & 0.81          & 0.80          & 0.80          & 0.60          & 0.57          & 0.56           \\ 
\hline
Tri-CRF                                                                    & 91\%                                                                                                 & 0.83          & 0.82          & 0.82          & 0.64          & 0.57          & 0.59           \\ 
\hline
Att-RNN                                                                    & 93\%                                                                                                 & 0.83          & 0.87          & 0.85          & 0.55          & 0.55          & 0.53           \\ 
\hline
Flaubert                                                                   & \textbf{94\%}                                                                                        & \textbf{0.89} & \textbf{0.91} & \textbf{0.90} & \textbf{0.69} & \textbf{0.74} & \textbf{0.70}  \\
\hline
\end{tabular}

%% file: crossVal.tex
\begin{tabular}{|c|c|c|c|c|c|c|} 
\hline
\multirow{2}{*}{\textbf{K \#}} & \multicolumn{3}{c|}{\textbf{Micro Average}}   & \multicolumn{3}{c|}{\textbf{Macro Average}}    \\ 
\cline{2-7}
                               & \textbf{P}    & \textbf{R}    & \textbf{F1}   & \textbf{P}    & \textbf{R}    & \textbf{F1}    \\ 
\hline
\textbf{1}                     & 0.93          & 0.93          & 0.93          & \textbf{0.79} & 0.75          & 0.75           \\ 
\hline
\textbf{2}                     & 0.93          & 0.94          & 0.94          & 0.76          & \textbf{0.80} & \textbf{0.77}  \\ 
\hline
\textbf{3}                     & 0.89          & 0.90          & 0.90          & 0.54          & 0.48          & 0.50           \\ 
\hline
\textbf{4}                     & 0.92          & 0.91          & 0.91          & 0.68          & 0.67          & 0.66           \\ 
\hline
\textbf{5}                     & \textbf{0.95} & \textbf{0.95} & \textbf{0.95} & 0.73          & 0.74          & 0.72           \\ 
\hhline{|=======|}
\textbf{avg}                   & 0.92          & 0.92          & 0.92          & 0.70          & 0.68          & 0.68           \\ 
\hline
\textbf{SD}                    & 0.02          & 0.02          & 0.02          & 0.10          & 0.12          & 0.10           \\
\hline
\end{tabular}

%% file: dialogueSqlite.tex
\begin{tabular}{ccccc} 
\hline
~\textbf{S/U} & ~\textbf{Type} & ~\textbf{Turn} & ~\textbf{Utterance}                                                                                                   & ~\textbf{TS}  \\ 
\hline
U             & ASR            & 1              & Paracétamol 1 g 4 times a day for 10 jours                                                                            & 40               \\
S             & TTS            & 1              & Choose the drug that matches your search                                                                              & 46               \\
U             & UI             & 2              & listview\_item\_clicked                                                                                               & 54               \\
S             & TTS            & 2              & Could you specify the dosage for the patient?                                                                         & 54               \\
U             & ASR            & 3              & 4 times a day for 2 weeks                                                                                             & 94               \\
S             & TTS            & 3              & \begin{tabular}[c]{@{}c@{}}4 times a day isn't precise enough. Please \\ give a more specific timeframe.\end{tabular} & 95               \\
U             & ASR            & 4              & 1 tablet at 8 in the morning, 1 at noon, 1 at 6pm and 1 at 10 pm                                                      & 138              \\
S             & ASR            & 4              & Do you confirm the this prescription?                                                                                 & 139              \\
U             & UI             & 5              & refuse\_button\_clicked                                                                                               & 150              \\
S             & TTS            & 5              & Click on the confirm button to cancel this prescription                                                               & 150              \\
\hline
\end{tabular}

%% file: metricsPxDial.tex
\begin{tabular}{|l|c|c|c|} 
\hline
\textbf{}                                                                                   & \textbf{Non-experts} & \textbf{Physicians} & \textbf{Other Experts}  \\ 
\hhline{|====|}
\textbf{Recordings}                                                                         & 1003                 & 605              & 459                       \\ 
\hline
\textbf{Task success rate}                                                                  & 86\%                 & 76\%             & 72\%                      \\ 
\hline
\textbf{AVG time per session}                                                               & 47.16 (s)            & 66.15 (s)        & 35.64 (s)                 \\ 
\hline
\textbf{AVG turn per session}                                                               & 3.5                  & 4.28             & 3.01                      \\ 
\hline
\textbf{AVG events per session}                                                             & 7.94                 & 8.44             & 6.4                       \\ 
\hline
\textbf{Nb of system crash}                                                                 & 4                    & 4                & 0                         \\ 
\hline
\begin{tabular}[c]{@{}c@{}}\textbf{Nb of system request}\\\textbf{for restart}\end{tabular} & 30                   & 32               & 11                        \\ 
\hline
\begin{tabular}[c]{@{}c@{}}\textbf{Nb of user request}\\\textbf{for restart}\end{tabular}   & 33                   & 29               & 8                         \\ 
\hline
\textbf{Nb of incomprehension}                                                              & 0.6\%                & 0.10\%           & 0.04\%                    \\ 
\hline
\textbf{Drug association rate}                                                              & 82\%                 & 87\%             & 95\%                      \\
\hline
\end{tabular}

%% file: PxDialStatsbyAge.tex
\begin{tabular}{|c|c|c|c|c|} 
\hline
\textbf{Age group}             & \textbf{18-28} & \textbf{29-40} & \textbf{41-59} & \textbf{60+}  \\ 
\hline
\textbf{Nb of Recordings}      & 674            & 578            & 377            & 283           \\ 
\hline
\textbf{Task Success Rate}     & 74\%           & 79\%           & 81\%           & 87\%          \\ 
\hline
\textbf{Avg Time Elapsed (s)}  & 37.02          & 42.92         & 65.79          & 64.49    \\ 
\hline
\textbf{Avg Dialogue Turns}    & 3.21           & 3.39           & 3.60           & 3.82          \\ 
\hline
\textbf{Avg nb of events}      & 6.87           & 7.33           & 8.15           & 9.06          \\ 
\hline
\textbf{Drug association rate} & 92\%           & 87\%           & 86\%           & 80\%          \\
\hline
\end{tabular}

%% file: statsDialSexe.tex
\begin{tabular}{|c|c|c|}
\hline
\textbf{}                                                                           & \textbf{Men} & \textbf{Women} \\ \hline
\textbf{\begin{tabular}[c]{@{}c@{}}Task Success Rate\end{tabular}}     & 76\%           & 80\%           \\ \hline
\textbf{Avg Time Elapsed (s)}                                                               & 51.99      & 43.62      \\ \hline
\textbf{Avg Dialogue Turns}                                                            & 3.36           & 3.41           \\ \hline
\textbf{Avg nb of events}                                                                 & 7.51           & 7.44           \\ \hline
\textbf{\begin{tabular}[c]{@{}c@{}}Drug association rate\end{tabular}} & 89\%           & 87\%           \\ \hline
\end{tabular}